\pdfoutput=1
\documentclass[11pt,letterpaper]{article}
\usepackage{emnlp2017}

\usepackage{times}
\usepackage{latexsym}
\usepackage{graphicx,accents}
\usepackage{tikz}
\usepackage{verbatim}
\usepackage{array}
\usepackage{multirow}

\newcolumntype{C}[1]{>{\centering\let\newline\\\arraybackslash\hspace{0pt}}m{#1}}

\makeatletter
\DeclareRobustCommand{\cev}[1]{%
	\mathpalette\do@cev{#1}%
}
\newcommand{\do@cev}[2]{%
	\fix@cev{#1}{+}%
	\reflectbox{$\m@th#1\vec{\reflectbox{$\fix@cev{#1}{-}\m@th#1#2\fix@cev{#1}{+}$}}$}%
	\fix@cev{#1}{-}%
}
\newcommand{\fix@cev}[2]{%
	\ifx#1\displaystyle
	\mkern#23mu
	\else
	\ifx#1\textstyle
	\mkern#23mu
	\else
	\ifx#1\scriptstyle
	\mkern#22mu
	\else
	\mkern#22mu
	\fi
	\fi
	\fi
}

\emnlpfinalcopy

\def\emnlppaperid{***}

\title{Mining fine-grained opinions on closed captions of YouTube videos with an attention-RNN}

\author{Edison Marrese-Taylor, Jorge A. Balazs, Yutaka Matsuo\\
	Graduate School of Engineering \\
	The University of Tokyo \\
	Tokyo, Japan \\
	emarrese,jorge,matsuo@weblab.t.u-tokyo.ac.jp}

\date{}

\begin{document}

\maketitle

\begin{abstract}
	Video reviews are the natural evolution of written product reviews. In this paper we target this phenomenon and introduce the first dataset created from closed captions of YouTube product review videos as well as a new attention-RNN model for aspect extraction and joint aspect extraction and sentiment classification. Our model provides state-of-the-art performance on aspect extraction without requiring the usage of hand-crafted features on the SemEval ABSA corpus, while it outperforms the baseline on the joint task. In our dataset, the attention-RNN model outperforms the baseline for both tasks, but we observe important performance drops for all models in comparison to SemEval. These results, as well as further experiments on domain adaptation for aspect extraction, suggest that differences between speech and written text, which have been discussed extensively in the literature, also extend to the domain of product reviews, where they are relevant for fine-grained opinion mining. 
\end{abstract}

\section{Introduction}

On-line videos have become indispensable to people's daily lives, as traffic statistics showed that by 2010 it accounted for 56.6\% of the total global consumer traffic \cite{siersdorfer2010howuseful}. Studies support the notion that on-line reviews can have a strong influence in the decision-making of potential Internet buyers \cite{chevalier2003effect}, thus becoming a major factor for both consumers and marketers \cite{hu2008online}. 

Video reviews are the natural evolution of written product reviews. In fact, people are increasingly turning to platforms such as YouTube to help them shop, looking for product reviews \cite{shoppingtrends2015}. YouTube unboxing videos have become a growing phenomenon \cite{shoppingtrends2015, youtubeinsights2014}. In 2015 alone, people in the U.S. watched 60M hours of them on YouTube, totaling 1.1 B views. The same year, views of product review videos increased by 40\% compared to 2014, and more than 1 million channels related to product reviews were counted \cite{youtubeviewers2015}. Despite all of this, the most widely used approaches in opinion mining focus only on tweets or written product reviews available on websites like Amazon. 

Therefore, in this paper we present the first opinion mining study focusing on \textit{video product reviews}. We take the fine-grained approach, which aims to detect the subjective expressions in text and to characterize their sentiment orientation, and analyze the closed captions of \textit{video product reviews} extracted from YouTube. Fine-grained opinion mining is important for a variety of NLP problems, including opinion-oriented question answering and opinion summarization, having been studied extensively in recent years. In practical terms, this approach defines the tasks of aspect extraction (\textit{AE}), sentiment classification (\textit{SC}) and a joint setting (\textit{AESC}). 

While \textit{AE} and \textit{AESC} have often been tackled as sequence labeling problem, where the sentence is a stream of tokens to be labeled using IOB and collapsed or sentiment-bearing IOB labels \cite{zhang_neural_2015} respectively, \textit{SC} can be regarded as a semantic compositional problem, where the obtained representation is used to predict the sentiment. 

Accounting for the patent differences between speech and written text, which have also led linguists to consider them as different domains \cite{biber1991variation} exhibiting different syntactic \cite{o1974syntactic} and distributional properties, we created the first annotated dataset using closed captions of YouTube product review videos, which we named the \textit{Youtubean} dataset.

Motivated by the success of attention-based approaches in multiple NLP problems such as machine translation \cite{bahdanau_neural_2015}, parsing \cite{vinyals_grammar_2015}, slot-filling \cite{liu_attention-based_2016} and others \cite{luong_effective_2015}, we also introduce an attention-augmented RNN model for \textit{AE} and \textit{AESC}. Compared to previous work, the attentional component makes our model specially suitable for \textit{AESC}, since it directly addresses the compositional nature of the sentiment classification task as it allows the model to represent the input sentence as a convex combination of word representations. This is confirmed by our results on the SemEval ABSA dataset \cite{pontiki_semeval_2014}, given that our model offers state-of-the-art performance for \textit{AESC} while also performing equivalently to the state-of-the-art for aspect extraction without the need for manually-crafted features.

We also show that our attention-RNN model outperforms the baseline for both \textit{AE} and \textit{AESC} on our dataset. However, we observed that compared to the SemEval corpora, all the tested models decreased their performance on it. As indicated by a descriptive analysis of our corpus and by additional experiments using domain adaptation techniques for \textit{AE}, which did not offer considerable gains, our results seem to support the existence of the aforementioned differences between speech and written text in the context of product reviews and their importance for fine-trained opinion mining. Our code and data are available for download on GitHub\footnote{\url{github.com/epochx/opinatt}}.

\section{Related Work}

Our work is related to aspect extraction using deep learning, a task that is often tackled as a sequence labeling problem. In particular, our work is related to \newcite{irsoy_opinion_2014}, who pioneered in the field by using multi-layered RNNs on a subset of the MPQA 1.2 dataset \cite{wiebe_annotating_2005}. Later, \newcite{liu_fine_grained_2015} successfully adapted the architectures by \newcite{mesnil_investigation_2013}, experimenting on the SemEval 2014 dataset \cite{pontiki_semeval_2014}. Compared to these, our model is novel since it introduces the usage of attention for \textit{AE}. In this sense, our work is also related to \newcite{liu_attention-based_2016}, who introduced an attention RNN for slot-filling in Natural Language Understanding.

We also find related work on the usage of RNNs for open domain targeted sentiment \cite{mitchell_open_2013}, where \newcite{zhang_neural_2015} experimented with neural CRF models using various RNN architectures on a dataset of informal language from Twitter. In our case, the domain is different since we focus on product reviews.

Regarding target-based sentiment analysis, we find several ad-hoc models that account for the sentence structure and the position of the aspect on it, such as \newcite{tang_aspect_2016} and \newcite{tang_effective_2016}, who use attention-augmented RNNs for the task. However, these models require the location of the aspect to be known in advance and therefore are only useful in pipeline models. Our work is similar to these since it also makes use of an attentional component to model compositionally in sentiment classification, but we model aspect extraction and sentiment classification as a joint task instead of using a pipeline approach.

\textit{AESC} has also often been tackled as a sequence labeling problem, mainly using CRFs \cite{mitchell_open_2013}. To model the problem in this fashion, collapsed or sentiment-bearing IOB labels \cite{zhang_neural_2015} are used. Pipeline models (i.e. task-independent model ensembles) have also been extensively studied by the same authors. We also find \newcite{xu_joint_2014} who performed \textit{AESC} by modeling the linking relation between aspects and the sentiment-bearing phrases.

When it comes to the video review domain, we find related work on YouTube mining, mainly focused on exploiting user comments. For example, \newcite{wu2014crowd} exploited crowdsourced texual data from time-synced commented videos, proposing a temporal topic model based on LDA. However, \newcite{dorner2013leave} showed that comments with references to video content\footnote{Class C7 in the paper} represent only 2\% to 4\% of comments in YouTube. Therefore, we think this kind of analysis might be limited. The work of \newcite{tahara2010nicoscene} introduced a similar approach for \textit{Nico Nico} using time-indexed social annotations to search for desirable scenes inside videos. 

\begin{table*}[h]
	\footnotesize
	\centering
	\begin{tabular}{c | c | c | c}
		\textbf{Video title} & \textbf{Video id} & \textbf{Length} & \textbf{\# of sentences} \\
		\hline
		Sprint Samsung Galaxy S5 Full Review! & jdzbw68mpZE & 10:23 & 97 \\
		Samsung Galaxy S5 Review & zV0u2UFwv6E & 12:07 & 147 \\
		Samsung Galaxy S5 Review - Phones 4u & 1lxAO\_YgZ98 & 5:07 & 41 \\
		Samsung Galaxy S5 Review & \_Ihe7jm63kU & 3:49 & 45 \\
		Samsung Galaxy S5 \textquotedblleft Special \textquotedblright Review \& Camera Samples & nayKYv\_7b6M & 12:00 & 52 \\
		Samsung Galaxy S5 vs Apple iPhone 5s: Which Is Better? & 1dvzHyHID0k & 3:34 & 32 \\
		Samsung Galaxy S5 review & bRv5JrKnp3M & 24:15 & 164 \\
	\end{tabular}
	\caption{Detail of the reviews used to create the \textit{Youtubean} dataset.}
	\label{table:youtubean_sources}
\end{table*}

On the other hand, \newcite{severyn_opinion_2014} proposed a systematic approach to mine user comments that relies on tree kernel models. Additionally, \newcite{krishna2013polarity} performed sentiment analysis on YouTube comments related to popular topics using machine learning techniques, showing that the trends in users' sentiments is well correlated to the corresponding real-world events. \newcite{siersdorfer2010howuseful} presented an analysis of dependencies between comments and comment ratings, proving that community feedback in combination with term features in comments can be used for automatically determining the community acceptance of comments. 

Finally, we find some papers that have successfully attempted to use closed caption mining for video activity recognition \cite{gupta2010usingclosed} and scene segmentation \cite{gupta2009usingclosed}. Similar work has been done using closed captions to classify movies by genre \cite{brezale2007using} and summarize video programs \cite{brezale2007using}.

\section{Dataset}

In YouTube, video authors con provide their own closed captions, or they can be generated automatically by the engine. In both cases, these captions can be interpreted as a time-indexed transcript of the speech in the video. Therefore, to minimize the amount of noise in the data, we utilized the user-provided closed captions of seven of the most popular reviews of the Samsung Galaxy S5 and creatd an annotated dataset for fine-grained opinion mining. We obtained, cleaned and processed the data, and annotated the aspects following the guidelines by \newcite{pontiki_semeval_2014} using \textit{brat}\footnote{\url{http://brat.nlplab.org/}} \cite{brat_annotation_2012}. We divided the annotation process into two steps. 

First, two different annotators tagged aspects independently, obtaining an exact inter-annotation agreement of 0.705 F1-score. This value rose to 0.823 when allowing for partial matches, which we defined as any overlap between the annotated terms. Discrepancies were discussed until a final setting was reached. 

With these annotations fixed, we asked the same annotators to tag the sentiment of each extracted aspect. On this task, the annotators obtained an average agreement of 0.942 F1-score. This time, discrepancies were discussed with a third person who acted as an arbiter, until an agreement was reached. Both aspect extraction and sentiment classification inter-annotator agreements are comparable to the values obtained in similar tasks \cite{jimenez_multi_lingual_2015} \cite{wiebe_annotating_2005}. 

\begin{table}[h!]
	\centering
	\footnotesize
	\begin{tabular}{c|c|c|c}
		\textbf{Corpus}         & \textbf{R}    & \textbf{L}    & \textbf{Y}      \\
		\hline
		\# Sentences            & 3041          & 3045          & 578             \\
		\# Aspects              & 1288          & 1042          & 525             \\ 
		Mean word/sentence      & 15.47         & 16.76         & 20.71           \\ 
		Mean const. tree depth  & 9.10          & 10.16         & 11.40           \\ 
		Mean word/aspect        & 1.97          & 1.83          & 2.14            \\ 
		Mean aspects/sentence   & 1.20          & 0.76          & 1.38            \\ 
		Sentences with aspects  & 66.46\%       & 48.87\%       & 66.96\%         \\
	\end{tabular}
	\caption{Descriptive corpora comparison.}
	\label{table:corpora_comparison}
\end{table}

Table \ref{table:youtubean_sources} provides some key information about the the source video reviews we have used to build our dataset, which we named the \textit{Youtubean} dataset. Table \ref{table:corpora_comparison} compares it to the SemEval Laptops and Restaurants corpora, regarded as the de facto datasets for written review mining. Several differences can be observed. A big distinction lies in mean sentence and aspect lengths, both of which are considerably longer in \textit{Youtubean}. We also analyzed sentence syntax complexity in terms of the constituency tree depth, observing that our sentence trees are deeper on average. Furthermore, \textit{Youtubean} exhibits both longer and more frequent aspect mentions.

\section{Proposed Model}

Our proposed model is a two-pass bidirectional RNN architecture that includes an attentional component. Formally, given an embedded input sequence $x = [ x_1, ..., x_n ]$ with one-hot encoded labels $y = [ y_1, ..., y_n ]$, we define the first pass as follows.
\begin{eqnarray}
\bar{x_i} = [ x_{i-d}; ...; x_{i}; ...; x_{i+d} ]\\
\vec h_i = \sigma (\bar{x_i}, \vec h_{i-1}) \\
\cev h_i = \sigma (\bar{x_i}, \cev h_{i+1}) \\
h_i =[ \vec h_i ;  \cev h_i]
\end{eqnarray}
Where $\sigma$ denotes the sigmoid nonlinearity, $\vec h_i$ and $\cev h_i$ are the forward and backward hidden states of the RNN, which are concatenated, and $\bar{x_i}$ is a context window of ordered word embedding vectors around position $i$, with a total size of $2d+1$. This context window is intended to improve the model capabilities to capture short-term temporal dependencies \cite{mesnil_investigation_2013}. 

The second pass goes through the hidden states $h_i$ and performs sequence labeling token by token. We use the attentional decoder from \cite{vinyals_grammar_2015}.
\begin{eqnarray}
u_{i,j} = v^{\top} \tanh(W_{\alpha} [h_i; h_j])\\
\alpha_{i,j} = softmax(u_{i,j}) \\
t_i = \sum_{j=1}^n \alpha_{i,j} \cdot h_j\\
\hat y_i = softmax(W_{s} [ h_i; t_i; y_{i-1}] )
\end{eqnarray}
Where $\hat y_i$ is a probability distribution over the label vocabulary for input $i$. As shown, this is obtained using both the corresponding \textit{aligned} input $h_i$ and the attention distribution over all hidden states $t_i$, i.e. using a global attention scheme \cite{luong_effective_2015}. While generating the output $\hat y_i$, we explicitly model the dependency on the previous label by adding $y_{i-1}$ to the computation. These two components are combined using a feed forward neural network, whose output dimension is the size of the tag label vocabulary for \textit{AE} or \textit{AESC}. To initialize the attention matrix $h_n$ is used so the model does not bypass it. As a loss function we use the mini-batch average cross-entropy.

The addition of the attentional component to our model is motivated by two factors. In the first place, in contrast to \newcite{mesnil_investigation_2013} who directly make use of a window of previous hidden states for AE, the attentional components allows us to access contextual information in a more natural and selective way. For AESC, the attention directly models sentiment compositionality.

\section{Experimental setup}

For our experiments, in addition to \textit{Youtubean}, we also worked with the SemEval ABSA 2014 Laptops and Restaurants corpora \cite{pontiki_semeval_2014}, which can be regarded as the de facto datasets for fine-grained review mining. For \textit{AE} we use the train/test splits provided for Phase B. For \textit{AESC}, since the test data does not have sentiment labels, we worked only with the training data. On the other hand, since the size of \textit{Youtubean} is smaller than the SemEval corpora, we used 5-fold cross validation to make results more robust. For each fold, we used 10\% of the development data as a validation set and compare our results using two-sided t-tests.

For evaluation, we used the CoNLL \textit{conlleval} script for evaluation based on F1-score. To perform joint aspect extraction and sentiment classification, we only considered \textit{positive} ($+$), \textit{negative} ($-$) and \textit{neutral} ($0$) as sentiment classes, and the additional \textit{conflict} class is mapped to \textit{neutral}. To gain insights on the output of the models for \textit{AESC}, we decoupled the IOB collapsed tags using simple heuristics to recover the \textit{simple} aspect extraction F1-score as well as classification performances for each sentiment class, but  we used the \textit{joint} tagging \textit{conlleval} F1-score to evaluate the models.

As a baseline, we implemented the RNN architectures by \newcite{liu_fine_grained_2015}, which are the state-of-the-art in fine-grained aspect extraction. We experimented with Jordan-style RNNs (JRNN), Elman-style RNNs (RNN), LSTMs and the bidirectional versions of these last two. We followed \newcite{irsoy_opinion_2014} to merge the forward and backward hidden states, setting $y_t = \sigma ( \vec U \vec h_t + \cev{U} \cev{h_t})$, where $\vec{U}$, $\cev{U}$ are output matrices for the forward and backward hidden states $\vec{h_t}$, $\cev{h_t}$, respectively. This gives the models more flexibility to capture complex relations in a sentence, making them able to learn how to weight future and past information.

For both our attention-RNN model and the baseline RNNs, we experimented with Senna embeddings \cite{collobert_natural_2011}, GoogleNews embeddings \cite{mikolov_distributed_2013} and WikiDeps \cite{levy_dependency_based_2014}. The usefulness of working with pre-trained embeddings for the baseline RNNs was already shown by \cite{liu_fine_grained_2015}. However, for comparison when experimenting with our model, we also used randomly initialized embeddings of sizes $50$ and $300$ to test this hypothesis.

To make our results more transparent, we explicitly experimented with two different pre-processing pipelines. We used Senna \cite{collobert_natural_2011}, which provides both a POS-tagger and a chunker, and CoreNLP \cite{manning_stanford_2014}. The latter lacks a chunker so we combined it with the CoNLL \textit{chunklink} script\footnote{http://ilk.uvt.nl/team/sabine/homepage/software.html}. As \newcite{liu_fine_grained_2015}, we also experimented adding the same 14 linguistic binary features they used, which are based on POS-tags and chunk IOB-tags. These are concatenated to the hidden layer of the RNN before the final output non-linearity. 

To train our baseline models we set a learning rate of $0.01$ with decay and early stopping on the validation set. We set a fixed window size of 1 for bi-directional and 3 for unidirectional models, and always train word embeddings. Exploratory experiments showed that most models stop learning after a few epochs ---3 or 4--- so we only trained for a maximum of 5 epochs. 

In the case of our attention-RNN model (ARNN), here we only report results using LSTMs, which outperformed all others cells we tried on preliminary experiments. We explored different hyper-parameter configurations, including context window sizes of $1$, $3$ and $5$ as well as hidden state sizes of $100$, $200$ and $300$, and dropout keep probabilities of $0.5$ and $0.8$. We also experimented concatenating the RNN hidden states after the first pass with the binary features used by \cite{liu_fine_grained_2015}. Finally, we also experimented with unidirectional versions of the RNNs. For training, we used mini-batch stochastic gradient descent with a mini-batch size of 16 and padded sequences to a maximum size of 200 tokens. We used exponential decay of ratio $0.9$ and early stopping on the validation when there was no improvement in the F1-score after 1000 training steps.

\section{Results}

\subsection{Aspect Extraction (\textit{AE})}

\subsubsection{Laptops}

Table \ref{table:baseline_results_laptops} summarizes our best baseline results on the Laptops datasets. For contrast we include the best F1-scores obtained by \newcite{liu_fine_grained_2015} (cf. F1* columns). We observed the CoreNLP pipeline outperformed the Senna pipeline, with an average absolute gain of $2.105\%$, significant at $p=1.29\times 10^{-5}$, and binary features proved useful offering average absolute gains of $1.538$\% ($p=1.29 \times 10^{-5}$). Finally, note that the best configurations always use SennaEmbeddings, which outperformed others significantly for each case.

\begin{table}[h!]
	\centering
	\footnotesize
	\begin{tabular}{c|c|c|c|c|c}
		\textbf{Model}   & \textbf{Emb.} & $|h|$ & \textbf{feat} & \textbf{F1}    & \textbf{F1*}      \\ \hline
		JRNN             & Senna         & 50    & No            & 70.81          & 73.42             \\ 
		LSTM             & Senna         & 100   & Yes           & 70.92          & \textbf{75.00}    \\ 
		BiLSTM           & Senna         & 50    & Yes           & 69.03          & 74.03             \\ 
		RNN              & Senna         & 50    & No            & \textbf{71.87} & 74.43             \\ 
		BiRNN            & Senna         & 50    & Yes           & 69.45          & 74.57             \\ 
	\end{tabular}
	\caption{Results of our implemented baseline RNN models on the Laptops dataset.}
	\label{table:baseline_results_laptops}
\end{table}

Table \ref{table:attention_rnn_results_laptops} summarizes the best results of our ARNN model on the Laptops dataset, where we obtained a maximum F1-score of $74.74$. Again, the CoreNLP pipeline significantly outperformed Senna, with an average absolute gain of $1.39$ ($p=3.4 \times 10^{-33}$) F1-score. Bidirectionality provided an absolute average gain of $1.15$ F1-score $(p=4.61 \times 10^-20)$. 

Both SennaEmbeddings and GoogleNews provided statistically equivalent results ($p=0.65$), which were also significantly superior to WikiDeps with p-values $9.54 \times 10^{17}$ and $2.6 \times 10^{-13}$ respectively. Pre-trained embeddings outperformed random embeddings on average, comparing across same-sized cases. Linguistic binary features did not statistically contribute to the performance.

\begin{table}[h!]
	\centering
	\footnotesize
	\begin{tabular}{l | c | c | c | c}
		\textbf{Embeddings} & $\mathbf{|d|}$ & $\mathbf{|cw|}$ & $\mathbf{|h|}$ & \textbf{F1}	\\ 
		\hline
		SennaEmbeddings 	& 50  & 1 & 100  & \textbf{74.74} 	\\
		Random 				& 50  & 3 & 300  & 70.19 			\\
		WikiDeps 	    	& 300 & 3 & 200  & 69.53     	    \\
		GoogleNews		    & 300 & 3 & 100  & 71.17     	    \\
		Random			    & 300 & 3 & 200  & 70.03     	    \\ 
	\end{tabular}
	\caption{Best results for our ARNN on \textit{AE} for the Laptops dataset.}
	\label{table:attention_rnn_results_laptops}
\end{table}

\subsubsection{Restaurants}

Table \ref{table:baseline_results_restaurants} summarizes our best baseline results for the Restaurants dataset, again for contrast we include the best F1-scores obtained by \newcite{liu_fine_grained_2015} (cf. F1* columns). 

Regarding the usage of the linguistic features, we found that they contributed to increasing performance with an average absolute gain of $1.083\%$ ($p=1.65 \times 10^{-6}$). This is consistent with previous findings by \newcite{liu_fine_grained_2015}. The Senna pipeline outperformed CoreNLP with an average absolute gain of $1.161\%$ ($p=1.02 \times 10^{-8}$). Embeddings caused statistically significant differences, where WikiDeps outperformed both other embeddings on average.

\begin{table}[h!]
	\centering
	\footnotesize
	\begin{tabular}{c|c|c|c|c|c}
		\textbf{Model}   & \textbf{Emb.} & $|h|$ & \textbf{feat} & \textbf{F1}    & \textbf{F1*}      \\ \hline
		JRNN             & WDeps         & 100   & Yes           & 78.20          & 79.89             \\ 
		LSTM             & WDeps         & 100   & Yes           & \textbf{78.97} & 81.37             \\ 
		BiLSTM           & WDeps         & 200   & Yes           & 74.73          & 81.06             \\ 
		RNN              & Senna         & 200   & Yes           & 77.13          & 81.66             \\ 
		BiRNN            & WDeps         & 100   & No            & 74.33          & \textbf{82.06}    \\ 
	\end{tabular}
	\caption{Results of our implemented baseline RNN models on the Restaurants dataset.}
	\label{table:baseline_results_restaurants}
\end{table}

Table \ref{table:attention_rnn_results_restaurants} summarizes the best results by our ARNN model on the Restaurants dataset, where we obtained a maximum F1-score of $81.83$. All of our best performing models use a bidirectional architecture. In fact, bidirectionality provided an average significant absolute gain of $0.89$ F1-score ($p=1.25 \times 10^{-17}$). Additionally, using CoreNLP as preprocessing pipeline provided an average gain of $0.585$ F1-score ($p=2.98 \times 10^{-21}$) over Senna.

\begin{table}[h!]
	\centering
	\footnotesize
	\begin{tabular}{l | c | c | c | c}
		\textbf{Embeddings} & $\mathbf{|d|}$ & $\mathbf{|cw|}$ & $\mathbf{|h|}$ & \textbf{F1}	\\ 
		\hline
		SennaEmbeddings 	& 50  & 1 & 100  & \textbf{81.83} 	\\
		Random 				& 50  & 3 & 100  & 78.79 			\\
		WikiDeps 	    	& 300 & 3 & 100  & 78.68     	    \\
		GoogleNews		    & 300 & 3 & 300  & 78.73     	    \\
		Random			    & 300 & 1 & 100  & 78.38     	    \\ 
		
	\end{tabular}
	\caption{Best results for our attention-RNNs on \textit{AE} on the Restaurants dataset.}
	\label{table:attention_rnn_results_restaurants}
\end{table}

Context windows proved beneficial as confirmed by the significantly different average F1-scores of $76.55$, $77.59$ and $77.28$ for window sizes 1, 3 and 5 respectively. We also observed significant performance differences using SennaEmbeddings, which outperformed all others with an average F1-score of $77.94$. GoogleNews and WikiDeps exhibited average F1-scores of $76.93$ and $76.55$, which are statistically different ($p=4.08 \times 10^{-6}$) and although they also outperformed random embeddings for $d=300$, they performed statistically worse than random embeddings for $d=50$. Linguistic binary features did not statistically contribute to the performance.

\subsubsection{\textit{Youtubean}}

Table \ref{table:baseline_youtubean_results} summarizes our results for baseline RNNs on \textit{Youtubean}. Again, we observed that adding linguistic features had a positive effect on the performance, with an average absolute gain of $1.30\%$ ($p=0.01$). SennaEmbeddings and WikiDeps provided better performance compared to GoogleNews, with average F1-scores of $49.11$, $49.64$ and $45.37$ respectively. The first two values were statistically indistinguishable. We could not observe significant differences in the performance for different pipelines. 

\begin{table}[h!]
	\centering
	\footnotesize
	\begin{tabular}{c|c|c|c|c|c}
		\textbf{RNN} & \textbf{Pipeline} & \textbf{Emb.} & \textbf{Feat.} & $\mathbf{|h|}$ & \textbf{F1}           \\ \hline
		RNN          & Senna         & WDeps        & Yes           & 100   & 55.82*                \\ 
		RNN          & CoreNLP       & WDeps        & No            & 200   & 55.69*                \\ 
		LSTM         & CoreNLP       & Senna        & No            & 100   & \textbf{56.13}        \\ 
		BiRNN         & CoreNLP       & WDeps        & No            & 200   & 50.15                 \\ 
		BiLSTM        & Senna         & Senna        & Yes           & 100   & 50.09                 \\ 
	\end{tabular}
	\caption{Results of our implemented baseline RNN models on \textit{AE} for the \textit{Youtubean} dataset.}
	\label{table:baseline_youtubean_results}
\end{table}

To further study the relation between written and video product reviews for aspect extraction, a task that has been broadly studied by our community, we complemented our RNNs baseline with two classic domain adaptation methods. Despite their simplicity, they are surprisingly difficult to beat \cite{daume_iii_domain_2006}. These techniques basically mean using each of the SemEval corpora as a source (SRC) dataset for transfer learning, where \textit{Youtubean} is set as the target (TGT). 

Our first domain-adaptation technique was WEIGHTED, a method that trains a model on the union of the SRC and TGT datasets, re-weighting examples from SRC \cite{daume_iii_domain_2006}. We did so by multiplying the input embedding matrix by the given weight $w$, which we set to $0.2$ based on the corpus size ratio. For training, we used 10-fold cross validation, adding all the examples of the SRC dataset to the training part of each fold-based arrangement. Since these model took longer to train we only experimented with the Senna pipeline. We omitted our bidirectional architectures given their poor performance and always included linguistic features, which generally contributed to an improved F1-score in our in-domain models. 

\begin{table}[h!]
	\footnotesize
	\centering
	\begin{tabular}{c | c | c | c | c}
		\textbf{RNN}    & \textbf{SRC} & \textbf{Emb.}     & $\mathbf{|h|}$  & \textbf{F1}    \\ \hline
		LSTM            & L            & Google            & 50          & 57.17          \\ 
		RNN             & L            & Google            & 100         & 55.12          \\ 
		JRNN            & L            & WDeps             & 200         & \textbf{58.30} \\ 
	\end{tabular}
	\caption{Results for the WEIGHTED technique.}
	\label{table:weighted_results}
\end{table}

As Table \ref{table:weighted_results} shows, using the Laptops dataset as SRC gives the best results in each case. Using this corpus led to an average absolute improvement over Restaurants of $3.79\%$ ($p=7.76 \times 10^{-11}$.) When it comes to embeddings, GoogleNews provided the best average performance with $53.44$ F1-score. However, this value was statistically indistinguishable at $p<0.08$ from WikiDeps, with an average $52.8$ F1-score.

Our second domain adaptation method was PRED, which uses the output of a SRC-trained classifier as a feature in the TGT model. Concretely, we first trained a model using all the examples on SRC. We then fed that model with all the TGT examples, adding its outputs as additional features to the TGT dataset, thus creating a new augmented version of it. Since these features are IOB-tags, we concatenate them with the linguistic features. We trained our models on the augmented TGT dataset, choosing the best performing settings from our previous experiments on \textit{AE}.

\begin{table}[h!]
	\centering
	\footnotesize
	\begin{tabular}{c|c|c|c|c}
		\textbf{RNN} & \textbf{SRC}     & \textbf{Emb.}     & $\mathbf{|h|}$   & \textbf{F1}       \\ \hline
		LSTM            & L             & Senna             & 100       & 56.83             \\  
		BiLSTM           & R             & WDeps             & 100       & 52.81             \\ 
		BiRNN            & R             & WDeps             & 100       & 52.99*            \\ 
		BiRNN            & R             & WDeps             & 200       & 52.90*            \\ 
		RNN             & R             & WDeps             & 100       & 57.70             \\ 
		JRNN            & R             & WDeps             & 200       & \textbf{59.69}    \\ 
	\end{tabular}
	\caption{Results for the PRED technique.}
	\label{table:pred_results}
\end{table}

Table \ref{table:pred_results} summarizes our results for PRED. We found that using Senna as the pre-processor provided better results in average, with an $0.89\%$ absolute gain significant at $p=0.01$. The Restaurants dataset provided better results than Laptops in average, with an absolute gain of $3.23\%$, significant at $p=8.78 \times 10^{-6}$.

Finally, Table \ref{table:attention_rnn_results_youtubean} shows our best results for our introduced ARNN in the \textit{Youtubean} dataset. For this case, we omitted random embeddings and binary features as previous experiments showed they did not contribute to increase the performance. 

\begin{table}[h!]
	\centering
	\footnotesize
	\begin{tabular}{c | c | c | c}
		\textbf{Embeddings} & $\mathbf{|cw|}$ & $\mathbf{|h|}$ & \textbf{F1}	\\ 
		\hline
		SennaEmbeddings 	& 3 & 100  & 56.28 	\\
		WikiDeps 	    	& 3 & 100  & 57.21     	    \\
		GoogleNews		    & 3 & 100  & \textbf{57.67}      	    \\
	\end{tabular}
	\caption{Best results for our ARNNs for \textit{AE} on \textit{Youtubean}.}
	\label{table:attention_rnn_results_youtubean}
\end{table}

\subsection{Joint aspect extraction and sentiment classification \textit{(AESC)}}

On our experiments for this task we based our parameter settings on the results for \textit{AE}, so we only used bidirectional ARNN models, and skipped binary features and random embeddings. 

\subsubsection{Laptops}

Table \ref{table:sent_results_laptops} summarizes our best results for the Laptops corpus. Based on the results for \textit{AE}, we only used CoreNLP as a pre-processing pipeline. For the RNN baseline, embeddings also reported significant differences, with SennaEmbeddings offering average absolute gains of $5.78$ F1-score ($p=10^{-4}$) over GoogleNews and $2.47$ F1-score ($p=8 \times 10^{-3}$) over WikiDeps.

For training our ARNN we only used the CoreNLP pipeline, since it significantly outperformed Senna in our experiments for \textit{AE}. All the values in the table were significantly different, although we observed different embeddings provided statistically equivalent results for certain lower performing parameter settings.

\begin{table}[h]
	\centering
	\scriptsize
	\begin{tabular}{c|c|cc|ccc}
		& & \multicolumn{2}{c}{\textbf{Tagging F1}} & \multicolumn{3}{|c}{\textbf{Classification F1}} \\
		\cline{3-7}
		\textbf{Model}       & \textbf{Emb.} & single    & joint     & $+$       & $-$       & $0$ \\
		\hline
		LSTM	    & Senna & 74.30	    		& \textbf{47.19} & 77.40     	   & 12.63     	    & 80.00 \\
		RNN     	& Senna & 74.08	    		& 46.52    		 & 77.13     	   & 17.70    		& \textbf{80.52} \\
		JRNN    	& Senna & \textbf{76.00}	& 46.62			 & \textbf{77.97}  & \textbf{22.86} & 80.39 \\
		\hline
		ARNN & Google    & 68.22         & 46.69          & 69.23      	& 62.69      	 & \textbf{86.83}   \\
		ARNN & Senna 	& \textbf{72.85}  & \textbf{52.46} & \textbf{73.23}	& \textbf{69.29} & 85.59       		\\
		ARNN & Wiki      & 71.46         & 50.85          & 63.94      	& 61.07       	 & 83.23       		\\
	\end{tabular}
	\caption{Results for \textit{AESC} on Laptops}
	\label{table:sent_results_laptops}
\end{table}


\subsubsection{Restaurants}

Regarding the Restaurants dataset, Table \ref{table:sent_results_restaurants} shows a summary of our best results. For this case, we only used the Senna pipeline, as it provided better results for \textit{AE}. We found that in the baseline RNNs SennaEmbeddings outperformed both other embeddings with average absolute gains of $2.37$ ($p=7.2 \times 10^{-4}$) and 3.36 ($p=1.19 \times 10^{-6}$) F1-score WikiDeps and GoogleNews, respectively.

For our ARNN, as in the previous case, we only used CoreNLP as preprocessing pipeline given that it provided better results for \textit{AE}. All the values in the table were significantly different.

\begin{table}[h]
	\centering
	\scriptsize
	\begin{tabular}{c|c|cc|ccc}
		& & \multicolumn{2}{c}{\textbf{Tagging F1}} & \multicolumn{3}{|c}{\textbf{Classification F1}} \\
		\cline{3-7}
		\textbf{Model}  & \textbf{Emb.} & single    & joint     & $+$       & $-$       & $0$ \\
		\hline
		LSTM  &	Senna & \textbf{69.24}    & \textbf{44.75}    & 67.81     		& \textbf{62.40}    & 87.22 \\
		RNN	  &	Senna  & 67.08     		& 40.64     		& \textbf{70.73}    & 58.47     		& \textbf{87.39} \\
		JRNN  &	Senna & 66.74     		& 40.65     		& 67.04     		& 49.29     		& 86.47 \\
		\hline
		ARNN & Google      & 73.80          & 50.63           & 78.90      	  & \textbf{53.25}	& 81.08 \\
		ARNN & Senna       & \textbf{79.57} & \textbf{54.75}  & 79.78      	  & 46.45			& 82.70 \\
		ARNN & Wiki        & 74.90          & 52.74           & \textbf{81.47}   & 51.39      		& \textbf{83.22} \\
	\end{tabular}
	\caption{Results for \textit{AESC} on Restaurants}
	\label{table:sent_results_restaurants}
\end{table}


\subsubsection{\textit{Youtubean}}

On \textit{Youtubean}, as Table \ref{table:sent_results_youtubean} shows, we see important performance drops compared to SemEval. In particular, the baseline models seem to be unable to correctly classify negative aspects. For this dataset, we found out that Senna provides better results than CoreNLP with an average absolute gain of $3.94$ F1-score, which was significant at $p=2.5\times 10^{-4}$. Embeddings did not provide statistically significant differences. Similarly, binary features did not statistically contribute to the performance either. 

\begin{table}[h]
	\centering
	\scriptsize
	\begin{tabular}{c|c|cc|ccc}
		& & \multicolumn{2}{c}{\textbf{Tagging F1}} & \multicolumn{3}{|c}{\textbf{Classification F1}} \\
		\cline{3-7}
		\textbf{Model}   & \textbf{Emb.} & single & joint  & $+$     & $-$   & $0$ \\
		\hline
		LSTM    & Senna & 41.32  		  & 25.38  		   & \textbf{35.83}  & \textbf{9.59} 	& 72.53 \\ 	 	
		RNN 	& Senna & \textbf{47.59}  & 30.12  		   & 0      		 & 0     			& \textbf{76.64} \\
		JRNN 	& Senna & 42.86  		  & \textbf{30.45}  & 23.33  		 & 0     			& 62.32 \\
		\hline
		ARNN & Google 	& 52.84 		 & 40.58 			& 45.39			 & \textbf{22.07} & 79.94 \\
		ARNN & Senna 	& 52.43			 & 41.17 			& 48.05 		 & 15.58 		  & 80.28 \\
		ARNN & Wiki 	& \textbf{55.50} & \textbf{41.49} 	& \textbf{52.32} & 14.85 		  & \textbf{81.07}
	\end{tabular}
	\caption{Results for \textit{AESC} on \textit{Youtubean}.}
	\label{table:sent_results_youtubean}
\end{table}



\section{Discussion}

Results for aspect extraction showed that our implemented RNN baseline performs similarly to the original models by \cite{liu_fine_grained_2015}, although we remained unable to replicate their exact numbers. Despite that, our attention-RNN is able to provide results that are better than our implementation and comparable to the original values for both Laptops and Restaurants datasets. Moreover, we achieved these results without the need to add the linguistic features, which did not offer significant performance differences in our experiments. We think the variable sentence representation introduced by the attentional component is able to model some of the semantics encoded in these binary features.

For aspect extraction in our dataset, we see our model is able to perform better than the baseline, again without the need to add manually-crafted features. However, simple domain adaptation techniques applied to the baseline RNNs managed to obtain the best results, adding a maximum of 3.56 F1-score over the baseline. We think this shows that video reviews and written reviews share some regularities, which could be exploited further to obtain better results. In this sense, it would be interesting to apply these domain adaptations techniques to our attention-RNN model and compare the results. However, regularities among these domains seem to be limited, given that our obtained gains were small and that no domain consistently delivered better performance.

Regarding \textit{AESC}, as shown by our decoupled results, we see all models slowly decreased their performance for aspect extraction, compared with results for \textit{AE}. This seems reasonable given the additional challenges of performing both tasks at the same time.

When it comes to sentiment classification, we see our attention-RNN outperforms the baseline RNNs by a solid margin. However, all models tend to perform poorly for the negative ($-$) class. We believe this may be related to the imbalanced nature of the datasets, or due to the additional composition challenges negation involves, which seem to be critical in our dataset. Compared to the baseline RNNs, which in some cases seemed basically unable to detect negative sentiment, our attention-RNN model offers increased, although yet limited capabilities to deal with the negative class. 

For \textit{AESC}, we also observed that SennaEmbeddings did not always provide top performances, being outperformed by other embeddings, even though the former were previously shown to offer the best performance for aspect extraction in all cases. We think this is related to the nature of the embeddings, since SennaEmbeddings were designed for the tasks in \cite{collobert_natural_2011} which do not include sentiment, while other embeddings can be regarded as general-purpose.

\section{Conclusions}

In this paper we presented the first fine-grained opinion mining study focusing on \textit{product video reviews}. We introduced the first annotated dataset for the domain, \textit{Youtubean}, and aspect extraction and \textit{AESC} with a novel attention-RNN. Our model offered state-of-the art performance for \textit{AESC} and results comparable to a strong RNN baseline for aspect extraction. Our descriptive corpus analysis as well as the performance obtained by all the models in our dataset suggest that differences between speech and written text, discussed extensively in the literature, also extend to the domain of product reviews, where they are relevant for fine-grained opinion mining. These findings introduce relevant research challenges and concrete paths for future researchers.

For future work, we plan to increase the size of our dataset and include reviews extracted from different product categories.  By doing this, we intend to make our results more robust and to further study the differences between written and video review, ultimately deriving new ways to overcome them. Finally, we also want to exploit the additional data from YouTube, such as the audio, video or specific frames extracted from it, and user comments, to improve our results. 

\bibliography{emnlp2017}
\bibliographystyle{emnlp_natbib}

\end{document}